\documentclass[letterpaper]{article} 
\usepackage{aaai2026}  
\usepackage{times}  
\usepackage{helvet}  
\usepackage{courier}  
\usepackage[hyphens]{url}  
\usepackage{graphicx} 
\usepackage{natbib}  
\usepackage{caption} 
\usepackage{algorithm}
\usepackage{algorithmic}

\usepackage{newfloat}
\usepackage{listings}

\usepackage{multirow}
\usepackage{bbding}
\usepackage{amsmath,amssymb,amsthm}
\usepackage{enumitem}
\usepackage[table,xcdraw]{xcolor}
\usepackage{booktabs}

\usepackage[most]{tcolorbox}

\usepackage{subcaption}

\definecolor{lightpink}{RGB}{255, 230, 230} 
\definecolor{darkred}{RGB}{185, 38, 74}   

\DeclareCaptionStyle{ruled}{labelfont=normalfont,labelsep=colon,strut=off} 
\floatstyle{ruled}
\newfloat{listing}{tb}{lst}{}
\floatname{listing}{Listing}
\theoremstyle{definition}                
\newtheorem{assumption}{Assumption}[section]   

\theoremstyle{plain}                     
\newtheorem{theorem}{Theorem}[section]
\newtheorem{lemma}[theorem]{Lemma}       
\newtheorem{proposition}[theorem]{Proposition}
\newtheorem{corollary}[theorem]{Corollary}

\theoremstyle{remark}                    

\DeclareMathOperator{\Var}{Var}

\title{Collaborative Representation Learning for Alignment of Tactile, Language, and Vision Modalities}
\author {
    Yiyun Zhou\textsuperscript{\rm 1}\equalcontrib,
    Mingjing Xu\textsuperscript{\rm 2}\equalcontrib,
    Jingwei Shi\textsuperscript{\rm 3}\equalcontrib,
    Quanjiang Li\textsuperscript{\rm 4},
    Jingyuan Chen\textsuperscript{\rm 1}\thanks{Corresponding author.}
}
\affiliations {
    \textsuperscript{\rm 1}Zhejiang University\\
    \textsuperscript{\rm 2}Swansea University\\
    \textsuperscript{\rm 3}Shanghai University of Finance and Economics\\
    \textsuperscript{\rm 4}National University of Defense Technology\\
    \{yiyunzhou, jingyuanchen\}@zju.edu.cn, mingjing.xu@swansea.ac.uk, shijingwei@stu.sufe.edu.cn,
    liquanjiang@nudt.edu.cn
}
\usepackage{bibentry}

\begin{document}

\maketitle

\begin{abstract}

Tactile sensing offers rich and complementary information to vision and language, enabling robots to perceive fine-grained object properties. However, existing tactile sensors lack standardization, leading to redundant features that hinder cross-sensor generalization. Moreover, existing methods fail to fully integrate the intermediate communication among tactile, language, and vision modalities. To address this, we propose TLV-CoRe, a CLIP-based Tactile-Language-Vision Collaborative Representation learning method. TLV-CoRe introduces a \textbf{\textit{Sensor-Aware Modulator} to unify tactile features across different sensors} and employs \textbf{tactile-irrelevant decoupled learning to disentangle irrelevant tactile features}. Additionally, a \textbf{\textit{Unified Bridging Adapter} is introduced to enhance tri-modal interaction within the shared representation space}. To fairly evaluate the effectiveness of tactile models, we further propose the RSS evaluation framework, focusing on Robustness, Synergy, and Stability across different methods. Experimental results demonstrate that TLV-CoRe significantly improves sensor-agnostic representation learning and cross-modal alignment, offering a new direction for multimodal tactile representation.

\end{abstract}

\section{Introduction}
\label{sec:introduction}

Tactile is one of the essential senses of human perception. Through tactile interaction, we can sense both static and dynamic attributes of objects (\textit{e.g.}, material texture, roughness, and hardness), many of which are too subtle to be reliably perceived by other perception systems like vision~\cite{cheng2025touch100k, shi2025presentagent, dave2024multimodal, li2025madakv, li2025flowmm, lv2025out, jiang2025acckv, jiang2025arnet}. In recent years, researchers have been striving to help robots understand the complex and realistic physical world by designing high-resolution tactile sensors~\cite{yuan2017gelsight, donlon2018gelslim, lambeta2020digit, zhang2024compact} comparable to human touch and collecting large-scale indoor and outdoor tactile image datasets~\cite{yang2022touch, kerr2022self, fu2024tou, yu2024octopi, feng2025anytouch}.

\begin{figure}[t]
    \includegraphics[width=\linewidth]{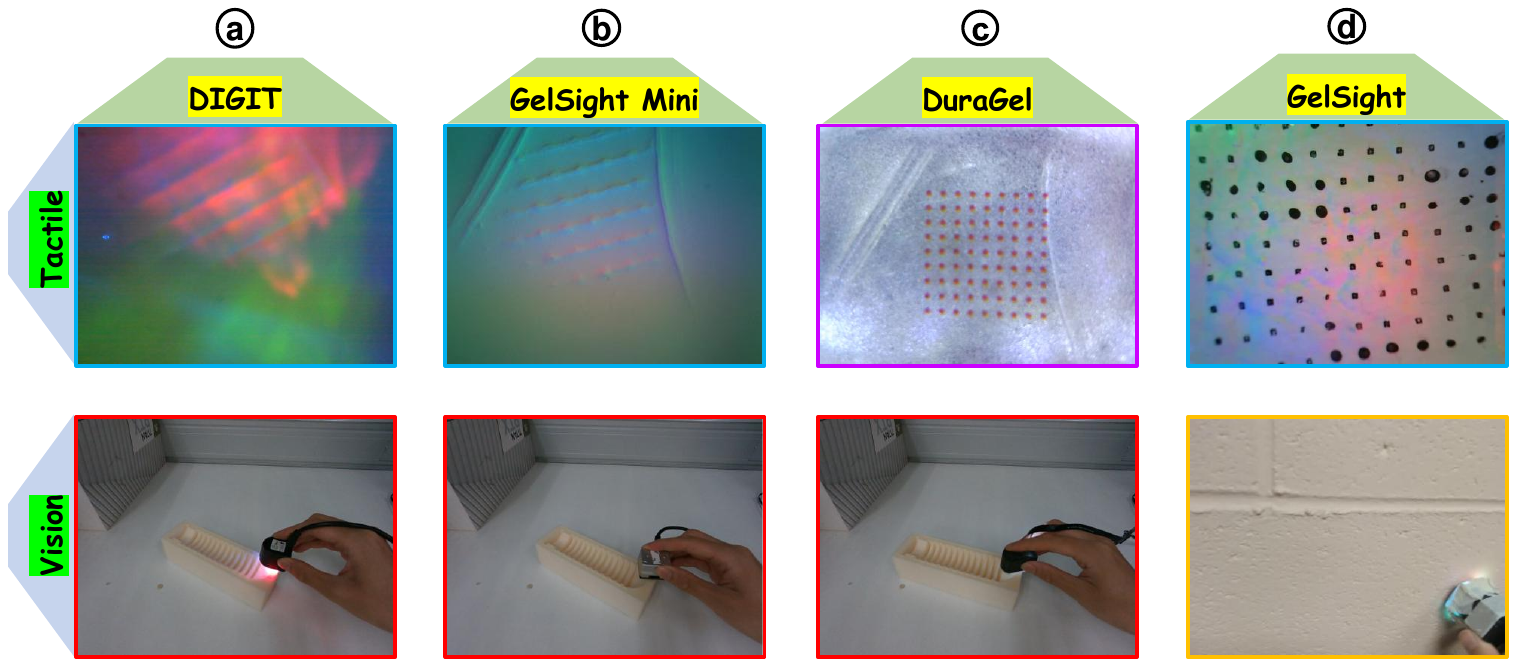}
    \vspace{-0.6cm}
    \caption{Three properties of heterogeneous sensors are identified: \texttt{(i)}. Tactile sensors lack full standardization, leading to significant tactile images variation~\cite{yang2024binding}. \texttt{(ii)}. Tactile images from the identical touch object can differ inconsistently (\textit{e.g.}, \textcircled{a} and \textcircled{b} are similar, both differing greatly from \textcircled{c}). \texttt{(iii)}. Despite different touch objects, tactile images may share a consistent style (\textit{e.g.}, \textcircled{a} and \textcircled{b} resemble \textcircled{d} in a dark tone tinged with red).
    \label{fig:sensor}}
    \vspace{-0.5cm}
\end{figure}

However, \textbf{tactile sensors are not yet fully standardized}. Due to external factors (\textit{e.g.}, camera type, lighting position, color, and illumination), tactile images can differ significantly even under identical touch object conditions (Fig.~\ref{fig:sensor}\texttt{(i)} and \texttt{(ii)}). To address these variations, previous studies~\cite{yang2024binding, feng2025anytouch} have borrowed the concept of positional encoding from language models~\cite{su2024roformer, zhao2024length, jiang2025purekv}, introducing learnable tokens to model sensor-specific characteristics. However, \textbf{these methods overlook a crucial fact: even when touch objects differ noticeably, the styles of the tactile images can still be quite similar} (Fig.~\ref{fig:sensor}\texttt{(iii)}), which poses a challenge for tactile models to disentangle tactile-irrelevant features.

Vision and language are also core channels for human-environment interaction. In real-world tasks, their integration has been extensively studied. A large body of work~\cite{radford2021learning, lv2025debiased, jia2021scaling, alayrac2022flamingo, li2023scaling} has successfully built semantic bridges between visual and linguistic modalities through contrastive learning~\cite{oord2018representation}, achieving remarkable progress. This success has since extended to additional modalities, including audio, point clouds, event etc~\cite{girdhar2023imagebind, guo2023point, wang2024omnibind, lyu2024unibind}. Despite the flourishing development of multimodal learning catalyzed by vision-language pretraining, \textbf{the tactile modality remains significantly underexplored}.

Recent research on tactile-language-vision learning has focused on representation learning based on CLIP~\cite{radford2021learning, cherti2023reproducible}. For instance, TLV-Link~\cite{cheng2025touch100k}, designed specifically for the GelSight sensor~\cite{yuan2017gelsight}, trains a tactile encoder via curriculum learning to achieve effective tri-modal alignment. AnyTouch~\cite{feng2025anytouch} proposes a unified representation learning framework for static-dynamic and multi-sensor tactile data, employing masked modeling, self-supervised multimodal alignment, and cross-sensor matching to improve generalization across different sensors. Vit-Lens-2~\cite{lei2024vit} introduces a generic multimodal encoding approach that first transforms various modality inputs into intermediate representations using lightweight modules and then feeds them into a frozen pretrained ViT~\cite{dosovitskiy2020image}, enabling efficient representation learning.
However, these methods face two challenges: (1) Most methods adapt LoRA~\cite{hu2022lora, zhou2025cola} within single modality branches, \textbf{without explicitly modeling the synergy among the three modalities before fusion, limiting their deep fusion capability}. (2) \textbf{There is a lack of standardized evaluation settings} (\textit{e.g.}, base models, batch sizes), making fair comparisons difficult.

To this end, we propose a CLIP-based \textbf{T}actile-\textbf{L}anguage-\textbf{V}ision \textbf{Co}llaborative \textbf{Re}presentation (\textbf{TLV-CoRe}) learning method, aiming to learn sensor-agnostic tactile representations. Unlike existing methods~\cite{yang2024binding, feng2025anytouch}, TLV-CoRe introduces a learnable \textit{Sensor-Aware Modulator} (SAM) that adaptively maps tactile features from different sensors into a unified parameter space, combined with tactile-irrelevant decoupled learning to effectively disentangle tactile-irrelevant features. Furthermore, to strengthen cross-modal collaboration in intermediate representations, TLV-CoRe incorporates a \textit{Unified Bridging Adapter} (UBA) into three modality encoders. UBA consists of modality-specific projection layers to learn individual representations and a shared feature space mapping layer to facilitate tri-modal alignment. 

For evaluation, we propose a \textbf{RSS} evaluation framework, designed to analyze the \textbf{R}obustness, \textbf{S}ynergy, and \textbf{S}tability of various tactile representation learning methods. We define three evaluation protocols—intra-sensor evaluation, cross-sensor generalization, and multi-sensor generalization—to assess the \textbf{robustness} of different methods. Given that \textbf{multimodal alignment should enhance rather than compromise individual modality performance}~\cite{wang2022deep, jiang2025fedcfa, wu2024semantic, li2025mergenet, dufumier2024align}, we introduce modal cross-evaluagtion tasks (especially between tactile and vision modalities) to assess \textbf{synergy} of various modal encoders. We also investigate the impact of batch size on model \textbf{stability}, as varying batch sizes affect the number of negative samples in contrastive learning. We encourage future research to adopt the RSS framework for comprehensive comparisons of different multimodal tactile methods based on CLIP.

Our key contributions are as follows:
\begin{itemize}[leftmargin=*, itemsep=0pt, topsep=0pt]
    \item{We design a \textit{Sensor-Aware Modulator} that enables flexible learning of unified tactile representations across multiple sensors and introduce tactile-irrelevant decoupled learning to effectively disentangle tactile-irrelevant features.}
    \item{We propose a novel \textit{Unified Bridging Adapter}, which includes separate projection layers for tactile, language, and vision encoders, as well as a shared projection to better align their representations.}
    \item{We provide a rigorous theoretical analysis of robustness, synergy and stability of our proposed method, providing valuable insights to guide the design of future tactile representation methods.}
    \item{We introduce a fair and comprehensive RSS evaluation framework to systematically analyze the robustness, synergy, and stability of other tactile representation learning methods, and verify the effectiveness of the proposed TLV-CoRe.}
\end{itemize}

Note that \textbf{the proposed RSS evaluation framework requires consistency in the base model and batch size to ensure that the evaluation focuses more on the differences in the design of the tactile representation methods}.

\section{Related Works}
\label{sec:related_works}

\begin{table*}[t!]
\centering
\begin{tabular}{@{}ccccc@{}}
\toprule[1.2pt]
Method     & Tactile    & Language     & Vision       & Base Model   \\ \midrule
TLV-Link~\cite{cheng2025touch100k}   & \Checkmark & \Checkmark   & \Checkmark   & CLIP-Based   \\
AnyTouch~\cite{feng2025anytouch}   & \Checkmark & \Checkmark   & \Checkmark   & CLIP-Based   \\
VIT-LENS-2~\cite{lei2024vit} & \Checkmark & \Checkmark   & \Checkmark   & CLIP-Based   \\
UniTouch~\cite{yang2024binding}   & \Checkmark & \Checkmark   & \Checkmark   & CLIP-Based / LLM-Based    \\
TVL-LLaMA~\cite{fu2024tou}  & \Checkmark & \Checkmark   & \Checkmark   & LLM-Based    \\
VT CMC~\cite{yang2022touch}     & \Checkmark & \XSolidBrush & \Checkmark   & Custom-Based \\
T3~\cite{zhao2024transferable}         & \Checkmark & \XSolidBrush & \XSolidBrush & Custom-Based \\
MViTac~\cite{dave2024multimodal}     & \Checkmark & \XSolidBrush & \Checkmark   & Custom-Based \\
TLA~\cite{hao2025tla}        & \Checkmark & \Checkmark   & \XSolidBrush & Custom-Based \\
SITR~\cite{gupta2025sensor}       & \Checkmark & \XSolidBrush & \XSolidBrush & Custom-Based \\ \bottomrule[1.2pt]
\end{tabular}%
\vspace{-0.2cm}
\caption{Summary of previous tactile representation learning methods in terms of tactile, language, vision, and base model.}\label{tab:prev_works}
\end{table*}

\subsection{Multimodal Alignment}
\label{sec:multimodal}
Multimodal alignment~\cite{baltruvsaitis2018multimodal,xu2023multimodal} aims to build bridges between different modalities. It not only helps models better understand cross-modal information, but indirectly facilitates representation learning within individual modalities~\cite{zhou2025revisiting, zhou2025disentangled, zhou2025cuff, zhang2024vision}. The success of CLIP~\cite{radford2021learning} has sparked rapid development in vision-language pretraining methods~\cite{jia2021scaling,alayrac2022flamingo,li2023scaling,zhao2023mmicl}. Subsequent research has extended alignment to additional modalities, \textit{e.g.}, audio~\cite{guzhov2022audioclip}, video~\cite{ma2022x}, and 3D point clouds~\cite{xue2023ulip}. ImageBind~\cite{girdhar2023imagebind} constructs a unified embedding space across six modalities through image-pairing learning, achieving impressive results in both visual and non-visual tasks. Inspired by ImageBind, Point-Bind~\cite{guo2023point} aligns 3D point clouds with 2D images, text, audio, and video by constructing a joint embedding space. UniBind~\cite{lyu2024unibind} further aligns multimodal embeddings to a large-model-enhanced embedding center via contrastive learning, resulting in a unified and balanced representation space. Most of these methods adopt the InfoNCE contrastive learning paradigm~\cite{oord2018representation}, and have demonstrated significant performance improvements, underscoring the effectiveness of this approach. Following this direction, we extend the concept of multimodal alignment to the relatively underexplored tactile modality, leveraging CLIP’s powerful vision-language pretraining capabilities.

\subsection{Tactile Representation Learning}
\label{sec:tactile}

In recent years, the large-scale collection of tactile datasets~\cite{yang2022touch, kerr2022self, fu2024tou, yu2024octopi, cheng2025touch100k, feng2025anytouch} and advances in tactile sensors~\cite{yuan2017gelsight, lambeta2020digit, zhang2024compact} have significantly accelerated research in tactile representation learning. VT CMC~\cite{yang2022touch} models tactile images from GelSight sensors~\cite{yuan2017gelsight} using contrastive multiview coding~\cite{tian2020contrastive}. T3~\cite{zhao2024transferable} proposes an architecture with sensor-specific encoders, a shared backbone network, and task-specific decoders, enabling transferable tactile representation learning across multiple sensors. MViTac~\cite{dave2024multimodal} uses an InfoNCE loss to jointly optimize visual and tactile features for effective intra- and inter-modal fusion. TLA~\cite{hao2025tla} encodes temporal tactile images into composite images and incorporates language reasoning for cross-modal finetuning, thereby promoting generalized tactile-language-action policy learning. SITR~\cite{gupta2025sensor} combines supervised contrastive learning with physics-based simulation to learn sensor-invariant representations, enabling zero-shot transfer across GelSight sensors. UniTouch~\cite{yang2024binding} aligns tactile signals with visual data and introduces learnable sensor-specific tokens to leverage CLIP for multimodal shared representations, supporting various zero-shot tactile tasks. Additionally, UniTouch integrates large language models to facilitate diverse tactile question-answering tasks. TVL-LLaMA~\cite{fu2024tou} proposes a tri-modal contrastive-trained tactile encoder aligned with vision and language, and further finetunes LLaMA2 to generate tactile descriptions from visual and tactile inputs. TLV-Link~\cite{cheng2025touch100k} combines teacher-student curriculum learning with contrastive learning for tactile-centric multimodal pretraining. AnyTouch~\cite{feng2025anytouch} integrates static and dynamic information through a hierarchical architecture, incorporating masked modeling, multimodal alignment, and cross-sensor matching for unified multi-sensor tactile representation learning. VIT-LENS-2~\cite{lei2024vit} leverages a pretrained ViT~\cite{dosovitskiy2020image} and modality-specific lens modules for efficient, scalable multimodal tactile learning.

As shown in Table~\ref{tab:prev_works}, the aforementioned methods fall into three categories: CLIP-Based, LLM-Based, and Custom-Based. The trend shows a shift towards custom end-to-end architectures, but this makes fair benchmarking difficult. Moreover, variations in batch size can significantly affect the stability of multimodal models on tactile tasks~\cite{higuera2024sparsh}. Our work focuses on CLIP-Based approaches, which allows us to use a consistent base model~\cite{cherti2023reproducible} and build a standardized RSS evaluation framework for fairer comparisons.

\begin{figure*}[t]
\includegraphics[width=\linewidth]{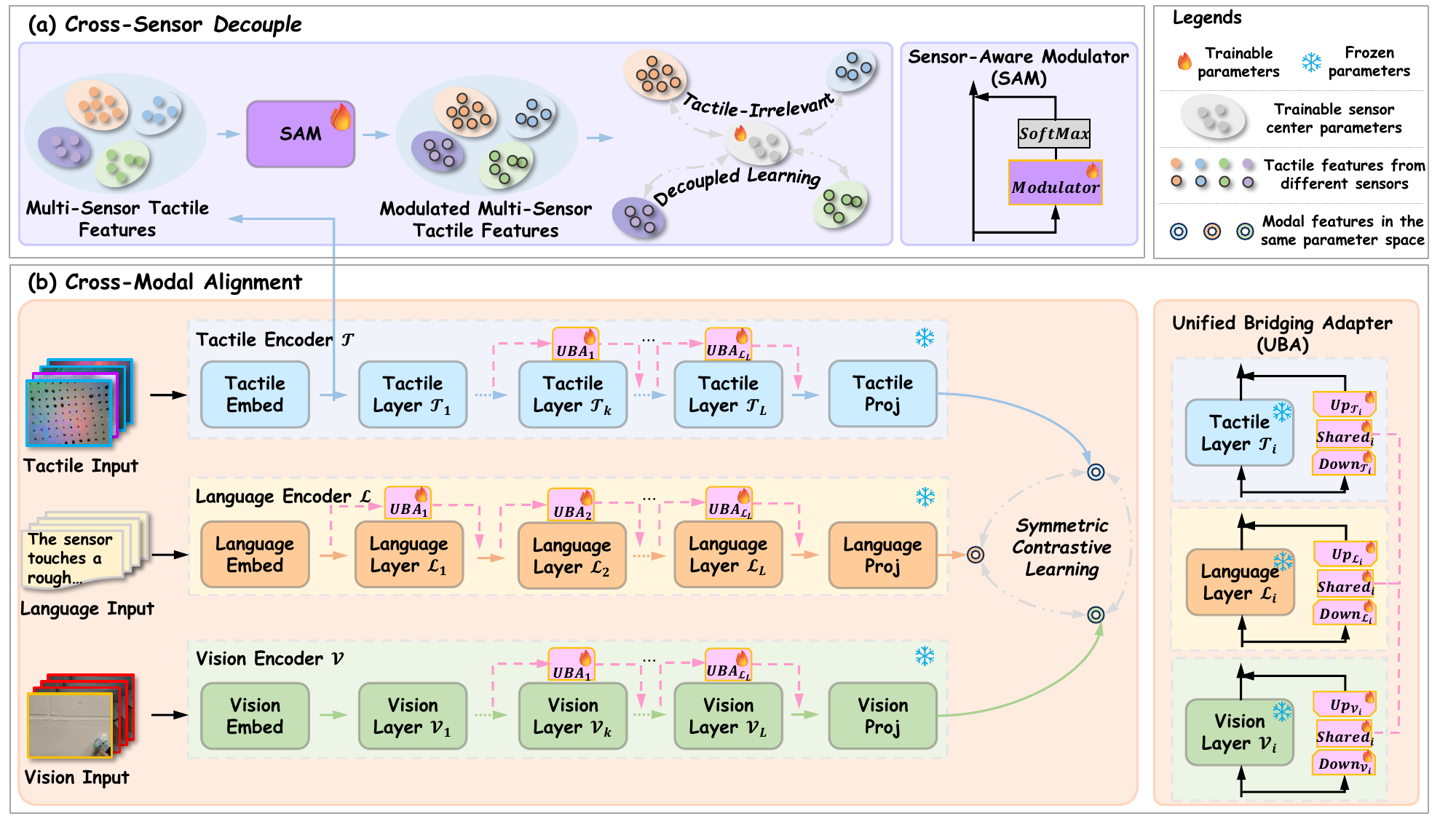}
\vspace{-0.6cm}
\caption{Overview of TLV-CoRe, which consists of modality-specific encoders for tactile (T), visual (V), and language (L) modalities inputs, a \textit{Sensor-Aware Modulator} (SAM) in the tactile branch to remove sensor-specific biases, and a \textit{Unified Bridging Adapter} (UBA) that projects features into a shared parameter space for alignment.}
\vspace{-0.5cm}
\label{fig:method}
\end{figure*}

\section{Methodology}
\label{sec:method}

\setlength{\abovedisplayskip}{3.5pt}
\setlength{\belowdisplayskip}{3.5pt}

We introduce TLV-CoRe, a method for learning collaborative representations that align the tactile (T), visual (V), and language (L) modalities in a shared latent space (see Fig.~\ref{fig:method}). TLV-CoRe comprises separate encoders for each modality and two modules: a \emph{Sensor-Aware Modulator} (SAM) in the tactile branch, and a cross-modal \emph{Unified Bridging Adapter} (UBA) that aligns the modalities. The SAM removes sensor-specific biases~\cite{zhao2024transferable,yang2024binding} from tactile features, yielding invariant representations across various tactile sensors. UBA is a lightweight module that projects features from each modality into a common latent space to facilitate alignment. We train the model with symmetric contrastive losses between modalities together with a sensor-invariance loss to enforce consistency across sensors. Finally, we provide theoretical guarantees of TLV-CoRe’s three properties. Detailed proofs can be found in the Appendix. We describe each component in detail below.

\subsection{Tactile Encoder and Sensor-Aware Modulator}
The tactile encoder $\mathcal{E}_T$ processes raw tactile inputs $x^T$ and produces a $d$-dimensional latent representation. Specifically, we implement $\mathcal{E}_T$ as a Vision Transformer (ViT)~\cite{dosovitskiy2020image}. The tactile image is divided into non-overlapping patches, each patch is linearly embedded, and the resulting sequence is processed through standard transformer blocks. This design enables the encoder to capture spatial patterns and tactile features.

To \textbf{address sensor-specific biases}, we incorporate a \emph{Sensor-Aware Modulator} (SAM) into the tactile branch. Let $s \in \{1,\dots,S\}$ index the sensor that captured a given sample. Given a tactile feature $h^T = \mathcal{E}_T(x^T)\in\mathbb{R}^d$, the SAM computes a routing-weight vector $r(h^T)\in\mathbb{R}^S$ via a learnable linear mapping and softmax: $r(h^T) = \mathrm{softmax}(W_r h^T)$, where $W_r \in \mathbb{R}^{S\times d}$. The $s$-th component $[r(h^T)]_s$ indicates the importance of sensor $s$. We then modulate the feature by 
\begin{equation}
h^T_{\mathrm{mod}} = h^T + [r(h^T)]_s\,h^T,
\end{equation}
which scales the feature according to the predicted sensor weight.

However, \textbf{tactile images captured under similar conditions can exhibit similar patterns across different sensors}~\cite{ou2024marker}. In such cases, the \textbf{SAM inadvertently clusters features by sensor identity rather than by underlying tactile content}. To address this, we employ \textbf{tactile-irrelevant decoupled learning that explicitly removes redundant information from the tactile representation}.

Specifically, we adversarially train the tactile encoder with a sensor-classification objective. We introduce a set of learnable sensor centroids $\{c_s\}_{s=1}^S\subset\mathbb{R}^d$. For a feature $h^T$, we compute its similarity to each centroid and define
\begin{equation}
p(s \mid h^T) = \frac{\exp\bigl(\langle h^T, c_s\rangle/\tau\bigr)}{\sum_{s'=1}^S \exp\bigl(\langle h^T, c_{s'}\rangle/\tau\bigr)},
\end{equation}
where $\langle \cdot, \cdot \rangle$ denotes cosine similarity and $\tau>0$ (\textit{e.g.}, 0.05) is a temperature. We then minimize the expected negative log-likelihood,
\begin{equation}
\mathcal{L}_{\text{DL}} = -\mathbb{E}_{(x^T,s)}\bigl[\log p(s\mid h^T)\bigr],
\end{equation}
and apply \textbf{a gradient reversal layer so that $\mathcal{E}_T$ learns to confuse the sensor classifier}. This adversarial training removes redundant information from $h^T$. \textbf{Combined with the SAM, this encourages the tactile features to capture intrinsic object properties rather than sensor artifacts}.

\subsection{Unified Bridging Adapter for Cross-Modal Alignment}
To \textbf{enable direct interactions between modalities}, we introduce a \emph{Unified Bridging Adapter} (UBA) in each branch. The UBA is a lightweight module that projects modality-specific features into a shared latent space. Formally, for each modality $m\in\{\mathrm{T},\mathrm{V},\mathrm{L}\}$, we define two projection matrices: $W^m_{\downarrow}\in\mathbb{R}^{r\times d}$ to down-project from $d$ to a lower dimension $r$ ($r\ll d$), and $W^m_{\uparrow}\in\mathbb{R}^{d\times r}$ to up-project back to $d$. We also introduce a shared transformation $W_{\mathrm{sh}}\in\mathbb{R}^{r\times r}$ common to all modalities: given a feature $h^m\in\mathbb{R}^d$ from modality $m$, we project it into the shared space, apply the shared transform, and project it back:
\begin{equation}
z^m_{\mathrm{shared}} = W_{\mathrm{sh}}\bigl(W^m_{\downarrow}h^m\bigr), 
\qquad
\Delta h^m = W^m_{\uparrow} z^m_{\mathrm{shared}}.
\end{equation}
We then form the aligned feature by adding this residual: $h^m_{\mathrm{aligned}} = h^m + \Delta h^m$.

Thus, $h^m_{\mathrm{aligned}}$ is a coordinated version of $h^m$ that has passed through the common latent space. By sharing $W_{\mathrm{sh}}$ across modalities, we ensure that the transformations at the bottleneck are identical for tactile, vision, and language streams, while the modality-specific matrices $W^m_{\downarrow}$ and $W^m_{\uparrow}$ allow each modality to interface with this shared space.

\paragraph{Practical UBA placement.} 
In our implementation we attach \emph{exactly} $L\!=\!12$ UBA blocks to \emph{each} modality so that every sample—regardless of its source—passes through the \textit{same number} of shared transformations before the final projection.  
Concretely, the language encoders comprise $12$ transformer layers; we attach one UBA per layer.  
The vision/tactile branch is deeper (CLIP design~\cite{cherti2023reproducible} with $24$ layers), so we leave \textbf{the first $12$ layers \emph{un‐bridged} to preserve low‑level visual/tactile primitives and attach UBAs only to the upper $12$ layers where high‑level semantics emerge}.

At layer $\ell$, the outputs of each modality’s transformer block are fed via its UBA and merged via residual addition. This multi-level UBA design ensures alignment occurs progressively at different semantic levels. After the final layer, we obtain aligned $h^T_{\mathrm{aligned}}, h^V_{\mathrm{aligned}}, h^L_{\mathrm{aligned}}$, which we $L_2$-normalize to yield final embeddings $z^T,z^V,z^L\in \mathbb{R}^d$ for cross-modal comparisons.

\subsection{Cross-Modal Contrastive Learning}
We train TLV-CoRe with symmetric contrastive losses for each pair of modalities and the sensor-invariance loss. For each pair $(X,Y)\in \{(\mathrm{T},\mathrm{V}),(\mathrm{T},\mathrm{L}),(\mathrm{V},\mathrm{L})\}$, we use a symmetric InfoNCE contrastive loss~\cite{oord2018representation}. Given a batch of $N$ aligned triplets $\{(z^T_i,z^V_i,z^L_i)\}_{i=1}^N$, the loss for the tactile--vision pair (as an example) is:

\begin{equation}
\begin{aligned}
&\mathcal{L}_{T\text{-}V} \;=\; -\frac{1}{2N}\sum_{i=1}^{N}\Bigl[
    \log\frac{\exp\bigl(\langle z^{T}_{i}, z^{V}_{i}\rangle/\tau\bigr)}
             {\sum_{j=1}^{N}\exp\bigl(\langle z^{T}_{i}, z^{V}_{j}\rangle/\tau\bigr)}
\\
&\hspace{1.6em}+\,\log\frac{\exp\bigl(\langle z^{V}_{i}, z^{T}_{i}\rangle/\tau\bigr)}
                  {\sum_{j=1}^{N}\exp\bigl(\langle z^{V}_{i}, z^{T}_{j}\rangle/\tau\bigr)}
\Bigr],
\end{aligned}
\end{equation}

where $\tau>0$ (\textit{e.g.}, 0.05) is a temperature parameter. The first term aligns each $z^T_i$ with its paired $z^V_i$, and the second term does the reverse, making the loss symmetric. We define analogous losses $\mathcal{L}_{T\text{-}L}$ and $\mathcal{L}_{V\text{-}L}$ for the tactile--language and vision--language pairs. The total alignment loss is 
$\mathcal{L}_{\text{SCL}} = \mathcal{L}_{T\text{-}V} + \mathcal{L}_{T\text{-}L} + \mathcal{L}_{V\text{-}L}$.

The overall training objective combines the alignment loss with the sensor-invariance loss:
\begin{equation}
\mathcal{L}_{\text{total}} = \mathcal{L}_{\text{SCL}} + \lambda_{\text{DL}}\,\mathcal{L}_{\text{DL}},
\end{equation}
where \textbf{$\lambda_{\text{DL}}>0$ balances the influence of sensor invariance}. We minimize $\mathcal{L}_{\text{total}}$ end-to-end (applying a gradient reversal layer to $\mathcal{L}_{\text{DL}}$). This objective trains the model to produce modality-agnostic representations: in the shared embedding space, matching tactile, visual, and textual inputs are embedded closely, while non-matching inputs are far apart. The sensor-invariance loss further ensures that tactile embeddings remain consistent across different sensors.

\subsection{Theoretical Guarantees}
\label{subsec:theory}

We analyze TLV-CoRe’s properties under standard optimization assumptions~\cite{patel2022global,xu2025psmgd,lei2019stochastic,pham2020proxsarah}. Let $\mathcal{L}(\Theta)$ denote the training objective. We make the following assumptions to facilitate analysis:

\begin{assumption}[Smoothness]
\label{assum:smoothness}
The gradient of $\mathcal{L}(\Theta)$ is $L$-Lipschitz continuous. Formally, for all parameters $\Theta, \Theta'$, $\|\nabla\mathcal{L}(\Theta) - \nabla\mathcal{L}(\Theta')\|
\;\le\; L\,\|\Theta - \Theta'\|$.

\end{assumption}

\begin{assumption}[Polyak--{\L}ojasiewicz (PL) Condition~\cite{karimi2016linear}]
\label{assum:pl}
In a neighborhood of a local optimum $\Theta^*$, the loss satisfies 
\[
\mathcal{L}(\Theta) - \mathcal{L}(\Theta^*)
\;\ge\; \frac{\mu}{2}\,\|\Theta - \Theta^*\|^2
\quad\text{for all }\Theta\text{ near }\Theta^*.
\]
\end{assumption}

\begin{assumption}[Bounded Gradient Variance]
\label{assum:bounded_variance}
The stochastic gradient has bounded variance. Specifically, $\mathbb{E}\bigl\|\nabla\mathcal{L}_{\mathcal{B}}(\Theta) - \nabla\mathcal{L}(\Theta)\bigr\|^2
\;\le\; \sigma^2$, 
where $\nabla\mathcal{L}_{\mathcal{B}}(\Theta)$ denotes the gradient on a mini-batch $\mathcal{B}$.
\end{assumption}

\subsubsection{Convergence Analysis}
\label{sec:conv1}

Under Assumptions~\ref{assum:smoothness}--\ref{assum:bounded_variance}, we obtain:

\begin{theorem}[Convergence Rate]
\label{thm:convergence}
Suppose Assumptions~\ref{assum:smoothness}--\ref{assum:bounded_variance} hold, and let $\Theta^*$ be a local minimizer satisfying the PL condition. Running SGD with step size $\eta<2/L$ gives:
\begin{equation}
\mathbb{E}\|\Theta_t - \Theta^*\|^2 
\;\le\; (1 - \eta\mu \beta)^t \|\Theta_0 - \Theta^*\|^2 
\;+\; \frac{\eta\sigma^2}{\mu\beta},
\end{equation}
where $\beta = 1/(1 + \kappa(W_{\mathrm{sh}}))$ and $\kappa(W_{\mathrm{sh}})$ is the condition number of the shared UBA matrix. 
\end{theorem}
Theorem~\ref{thm:convergence} shows that sharing the UBA across modalities accelerates convergence by improving the effective condition number.

\begin{table*}[t]
\centering
\begin{tabular}{@{}c|ccc|c|c@{}}
\toprule[1.2pt]
\textbf{Dataset}                & \textbf{Tactile}                   & \textbf{Language}                  & \textbf{Vision}                    & \textbf{Sensor}                                  & \textbf{Size} \\ \midrule
The Feeling of Success & \Checkmark &      \XSolidBrush                     & \Checkmark & GelSight                                & 9.3k \\
ObjectFolder 1.0       & \Checkmark &        \XSolidBrush                   & \Checkmark & DIGIT                                   & 8.3k \\
ObjectFolder 2.0       & \Checkmark &       \XSolidBrush                    & \Checkmark & GelSight                                & 7.2k \\
Touch and Go           & \Checkmark & \Checkmark & \Checkmark & GelSight                                & 250k \\
SSVTP                  & \Checkmark & \Checkmark & \Checkmark & DIGIT                                   & 4.5k \\
TVL                   & \Checkmark & \Checkmark & \Checkmark & DIGIT                                   & 39k  \\
Octopi                 & \Checkmark & \Checkmark &   \XSolidBrush                        & GelSight Mini                           & 39k  \\
TacQuad                & \Checkmark & \Checkmark & \Checkmark & GelSight, DIGIT, DuraGel, GelSight Mini & 55k  \\ \bottomrule[1.2pt]
\end{tabular}
\vspace{-0.2cm}
\caption{Statistics of the datasets used in the experiment.}\label{tab:statics}
\end{table*}

\subsubsection{Robustness via Sensor-Invariance}

We next examine how removing sensor-specific information via the SAM affects training robustness.

\begin{lemma}[Gradient Variance Reduction]
\label{lemma:variance}
Let $\mathcal{I}(h^T; s)$ denote the mutual information between the tactile representation $h^T$ and the sensor identity $s$. As the model removes sensor-specific features, the variance of the stochastic gradient satisfies 
\begin{equation}
\mathrm{Var}[\nabla \mathcal{L}(\Theta)] \;\le\; \sigma_0^2 \;-\; \gamma\,\mathcal{I}(h^T; s),
\end{equation}
for constants $\sigma_0^2,\gamma>0$. Hence, reducing $\mathcal{I}(h^T; s)$ (via the SAM) lowers the gradient variance.
\end{lemma}

\begin{proposition}[Optimization Robustness]
\label{prop:robustness}
Since the SAM drives $\mathcal{I}(h^T; s)\to \varepsilon$ (with $\varepsilon\ge0$ small), the asymptotic gradient variance is bounded by 
\begin{equation}
\limsup_{t\to\infty}\mathrm{Var}[\nabla \mathcal{L}(\Theta_t)] 
\;\le\; \sigma_0^2 - \gamma(1-\varepsilon).
\end{equation}
Thus, as sensor-specific information is eliminated, the training gradients become more stable.
\end{proposition}

Lemma~\ref{lemma:variance} and Proposition~\ref{prop:robustness} explain ours training robustness: removing sensor-specific signals via the SAM reduces stochastic gradient noise and leads to more stable optimization.

\subsubsection{Cross-Modal Synergy}

The UBA also enables information transfer across modalities, under the following assumption:

\begin{assumption}[Shared and Unique Information]
\label{assum:information}
Each modality $m\in\{T,V,L\}$ encodes information about the task label $Y$, with components unique to $m$ and components shared across modalities.
\end{assumption}

\begin{theorem}[Cross-Modal Information Transfer]
\label{thm:cross_modal}
Under Assumption~\ref{assum:information}, aligning modality $m$ with modality $m'$ via the UBA increases its label information. Formally,

\begin{equation}
\begin{aligned}
&\mathcal{I}\!\bigl(h^m_{\mathrm{aligned}};Y\bigr)
\;\ge\;\mathcal{I}\!\bigl(h^m;Y\bigr)
\\
&+\;\alpha\,\min\!\bigl\{\,r,\;\mathcal{I}\!\bigl(h^{m'};Y\bigr)-\mathcal{I}\!\bigl(h^m;Y\bigr)\bigr\},
\end{aligned}
\end{equation}
where $r$ is the dimension of the UBA’s shared subspace and $\alpha\in(0,1)$ is a constant. Thus, $h^m_{\mathrm{aligned}}$ can gain up to $\alpha r$ bits of information that modality $m'$ has but $m$ lacks.
\end{theorem}

\begin{table*}[ht!]
\centering
\setlength{\tabcolsep}{1.3mm}
\begin{tabular}{@{}c|c|c|ccc|c|c|
c@{}}
\toprule[1.2pt]
\multirow{4}{*}{Training Data} & \multirow{3}{*}{Method} & \multirow{3}{*}{\%Param} &         & TAG &          & OF 1.0 & OF 2.0 & \multicolumn{1}{c}{Feel}  \\ \cmidrule(lr){4-6} \cmidrule(lr){7-7} \cmidrule(lr){8-8} \cmidrule(lr){9-9}
                           &    &                         & Material & Roughness    & Hardness & Material                                  & Material & \multicolumn{1}{c}{Grasp}                                 \\ \cmidrule(l){2-9} 
                               & CLIP~\cite{cherti2023reproducible}   &       -          &     {\cellcolor[HTML]{F0F0F0}52.73}     &     {\cellcolor[HTML]{F0F0F0}82.16}         &    {\cellcolor[HTML]{F0F0F0}85.32}       &                    {\cellcolor[HTML]{F0F0F0}41.15}              &          {\cellcolor[HTML]{F0F0F0}72.97}  &    \multicolumn{1}{c}{{\cellcolor[HTML]{F0F0F0}72.52}}                   \\ \midrule
\multirow{4}{*}{TAG}           & TLV-Link\dag~\cite{cheng2025touch100k}   &     1.23        &    {\cellcolor[HTML]{F6E6E6}53.26}      &       {\cellcolor[HTML]{F6E6E6}84.80}       &     {\cellcolor[HTML]{F6E6E6}85.94}     &      {\cellcolor[HTML]{E2F0D9}43.75}                             &        {\cellcolor[HTML]{E2F0D9}74.12}        & \multicolumn{1}{c}{{\cellcolor[HTML]{E2F0D9}76.01}}                 \\
                               & AnyTouch~\cite{feng2025anytouch}   &   1.31          &     {\cellcolor[HTML]{F6E6E6}61.48}     &       {\cellcolor[HTML]{F6E6E6}86.31}       &    {\cellcolor[HTML]{F6E6E6}85.32}      &     {\cellcolor[HTML]{E2F0D9}43.88}                              & {\cellcolor[HTML]{E2F0D9}75.20}  & \multicolumn{1}{c}{{\cellcolor[HTML]{E2F0D9}80.53}}                                \\
                               & VIT-LENS-2~\cite{lei2024vit}  &    7.00        &   {\cellcolor[HTML]{F6E6E6}\textbf{65.99}}       &   {\cellcolor[HTML]{F6E6E6}87.16}       &    {\cellcolor[HTML]{F6E6E6}91.08}      &         {\cellcolor[HTML]{E2F0D9}37.00}                          &      {\cellcolor[HTML]{E2F0D9}75.85}       & \multicolumn{1}{c}{{\cellcolor[HTML]{E2F0D9}-}}                      \\
                               & TLV-CoRe   &     0.30        &      {\cellcolor[HTML]{F6E6E6}65.44}    &           {\cellcolor[HTML]{F6E6E6}\textbf{88.81}}   &   {\cellcolor[HTML]{F6E6E6}\textbf{92.65}}       &   {\cellcolor[HTML]{E2F0D9}\textbf{49.12}}                                &          {\cellcolor[HTML]{E2F0D9}\textbf{76.28}}        & \multicolumn{1}{c}{{\cellcolor[HTML]{E2F0D9}\textbf{81.28}}}                 \\ \midrule
\multirow{4}{*}{SSVTP}         & TLV-Link   &  1.23           &    {\cellcolor[HTML]{E2F0D9}55.52}      &        {\cellcolor[HTML]{E2F0D9}84.63}      &    {\cellcolor[HTML]{E2F0D9}86.32}      &      {\cellcolor[HTML]{E2F0D9}36.38}                             &         {\cellcolor[HTML]{E2F0D9}75.45}      & \multicolumn{1}{c}{{\cellcolor[HTML]{E2F0D9}74.88}}                    \\
                               & AnyTouch    & 1.31           &    {\cellcolor[HTML]{E2F0D9}62.49}      &     {\cellcolor[HTML]{E2F0D9}67.19}        &    {\cellcolor[HTML]{E2F0D9}73.93}       &          {\cellcolor[HTML]{E2F0D9}40.12}                         &  {\cellcolor[HTML]{E2F0D9}71.46}   & \multicolumn{1}{c}{{\cellcolor[HTML]{E2F0D9}68.26}}                              \\
                               & VIT-LENS-2   &  7.00         &     {\cellcolor[HTML]{E2F0D9}48.95}     &       {\cellcolor[HTML]{E2F0D9}\textbf{86.91}}       &    {\cellcolor[HTML]{E2F0D9}83.75}      &                         {\cellcolor[HTML]{E2F0D9}35.38}          &          {\cellcolor[HTML]{E2F0D9}75.00} & \multicolumn{1}{c}{{\cellcolor[HTML]{E2F0D9}-}}                         \\
                               & TLV-CoRe    &  0.30          &   {\cellcolor[HTML]{E2F0D9}\textbf{63.25}}       &    {\cellcolor[HTML]{E2F0D9}85.39}          &   {\cellcolor[HTML]{E2F0D9}\textbf{86.78}}       &     {\cellcolor[HTML]{E2F0D9}\textbf{48.50}}                              &    {\cellcolor[HTML]{E2F0D9}\textbf{75.74}}     & \multicolumn{1}{c}{{\cellcolor[HTML]{E2F0D9}\textbf{75.39}}}                          \\ \midrule
\multirow{4}{*}{TVL}           & TLV-Link    &  1.23         &   {\cellcolor[HTML]{E2F0D9}51.14}        &   {\cellcolor[HTML]{E2F0D9}80.00}            &     {\cellcolor[HTML]{E2F0D9}84.33}      &                  {\cellcolor[HTML]{E2F0D9}40.50}                 &     {\cellcolor[HTML]{E2F0D9}75.38}          & \multicolumn{1}{c}{{\cellcolor[HTML]{E2F0D9}76.06}}                   \\
                               & AnyTouch    &   1.31         &   {\cellcolor[HTML]{E2F0D9}46.18}       &      {\cellcolor[HTML]{E2F0D9}84.39}         &    {\cellcolor[HTML]{E2F0D9}73.74}      &         {\cellcolor[HTML]{E2F0D9}41.88}     &       {\cellcolor[HTML]{E2F0D9}75.41}     &  \multicolumn{1}{c}{{\cellcolor[HTML]{E2F0D9}77.57}}                       \\
                               & VIT-LENS-2   &  7.00         &   {\cellcolor[HTML]{E2F0D9}52.64}       &    {\cellcolor[HTML]{E2F0D9}82.16}          &  {\cellcolor[HTML]{E2F0D9}80.53}        &    {\cellcolor[HTML]{E2F0D9}38.26}                               &    {\cellcolor[HTML]{E2F0D9}76.24} & \multicolumn{1}{c}{{\cellcolor[HTML]{E2F0D9}-}}                               \\
                               & TLV-CoRe     &  0.30         &  {\cellcolor[HTML]{E2F0D9}\textbf{54.47}}        &           {\cellcolor[HTML]{E2F0D9}\textbf{84.54}}   &    {\cellcolor[HTML]{E2F0D9}\textbf{84.47}}      &     {\cellcolor[HTML]{E2F0D9}\textbf{45.13}}                              &  {\cellcolor[HTML]{E2F0D9}\textbf{77.89}}             & \multicolumn{1}{c}{{\cellcolor[HTML]{E2F0D9}\textbf{77.95}}}                    \\ \midrule

\multirow{4}{*}{Octopi$^\ast$}        & TLV-Link     &   1.23        &     {\cellcolor[HTML]{E2F0D9}48.72}     &       {\cellcolor[HTML]{E2F0D9}79.55}       &     {\cellcolor[HTML]{E2F0D9}81.97}     &          {\cellcolor[HTML]{E2F0D9}47.12}                         &     {\cellcolor[HTML]{E2F0D9}73.58}  & \multicolumn{1}{c}{{\cellcolor[HTML]{E2F0D9}74.99}}                             \\
                               & AnyTouch     &  1.31         &    {\cellcolor[HTML]{E2F0D9}44.39}     &      {\cellcolor[HTML]{E2F0D9}\textbf{86.36}}        &     {\cellcolor[HTML]{E2F0D9}81.13}    &  {\cellcolor[HTML]{E2F0D9}38.12}                                 &     {\cellcolor[HTML]{E2F0D9}73.11}       & \multicolumn{1}{c}{{\cellcolor[HTML]{E2F0D9}79.07}}                       \\
                               & VIT-LENS-2   &  7.00         &     {\cellcolor[HTML]{E2F0D9}48.11}     &      {\cellcolor[HTML]{E2F0D9}82.02}        &     {\cellcolor[HTML]{E2F0D9}84.36}     &    {\cellcolor[HTML]{E2F0D9}39.62}            &        {\cellcolor[HTML]{E2F0D9}75.13}           &    \multicolumn{1}{c}{{\cellcolor[HTML]{E2F0D9}-}}                               \\
                               & TLV-CoRe     &  0.30         &       {\cellcolor[HTML]{E2F0D9}\textbf{52.65}}   &     {\cellcolor[HTML]{E2F0D9}85.83}         &   {\cellcolor[HTML]{E2F0D9}\textbf{86.43}}       &    {\cellcolor[HTML]{E2F0D9}\textbf{48.88}}                               &  {\cellcolor[HTML]{E2F0D9}\textbf{75.86}}                          &    \multicolumn{1}{c}{{\cellcolor[HTML]{E2F0D9}\textbf{80.63}}}   \\ \midrule
\multirow{4}{*}{TacQuad}                          & TLV-Link  &  1.23            &   {\cellcolor[HTML]{E8F1F9}56.60}        &       {\cellcolor[HTML]{E8F1F9}83.93}       &    {\cellcolor[HTML]{E8F1F9}87.37}       &          {\cellcolor[HTML]{E8F1F9}37.25} &           {\cellcolor[HTML]{E8F1F9}\textbf{76.53}}     &    \multicolumn{1}{c}{{\cellcolor[HTML]{E8F1F9}76.12}}              \\
                                                  & AnyTouch  &   1.31           &  {\cellcolor[HTML]{E8F1F9}45.14}      &     {\cellcolor[HTML]{E8F1F9}84.12}         &   {\cellcolor[HTML]{E8F1F9}80.61}       &             {\cellcolor[HTML]{E8F1F9}41.62}                      &                    {\cellcolor[HTML]{E8F1F9}74.29}     & \multicolumn{1}{c}{{\cellcolor[HTML]{E8F1F9}80.42}}     \\
                                                  & VIT-LENS-2   &    7.00       &     {\cellcolor[HTML]{E8F1F9}47.50}     &    {\cellcolor[HTML]{E8F1F9}85.94}          &     {\cellcolor[HTML]{E8F1F9}84.44}     &      {\cellcolor[HTML]{E8F1F9}39.62}                             &     {\cellcolor[HTML]{E8F1F9}75.25} & \multicolumn{1}{c}{{\cellcolor[HTML]{E8F1F9}-}}                              \\
                                                  & TLV-CoRe    &    0.30        &    {\cellcolor[HTML]{E8F1F9}\textbf{58.37}}      &   {\cellcolor[HTML]{E8F1F9}\textbf{86.80}}           &    {\cellcolor[HTML]{E8F1F9}\textbf{87.52}}     &    {\cellcolor[HTML]{E8F1F9}\textbf{42.25}}                                &   {\cellcolor[HTML]{E8F1F9}75.91}   & \multicolumn{1}{c}{{\cellcolor[HTML]{E8F1F9}\textbf{80.77}}}                             \\ \midrule
TAG,SSVTP, & TLV-Link      &    1.23      &    {\cellcolor[HTML]{E8F1F9}54.82}      &    {\cellcolor[HTML]{E8F1F9}84.53}          &    {\cellcolor[HTML]{E8F1F9}86.78}      &              {\cellcolor[HTML]{E8F1F9}42.64}                     &        {\cellcolor[HTML]{E8F1F9}75.58}      & \multicolumn{1}{c}{{\cellcolor[HTML]{E8F1F9}76.39}}                     \\
                                        Octopi,          & AnyTouch      &  1.31        &   {\cellcolor[HTML]{E8F1F9}56.43}       &     {\cellcolor[HTML]{E8F1F9}85.72}         &  {\cellcolor[HTML]{E8F1F9}84.31}        &     {\cellcolor[HTML]{E8F1F9}44.12}                              &        {\cellcolor[HTML]{E8F1F9}76.50} & \multicolumn{1}{c}{{\cellcolor[HTML]{E8F1F9}79.24}}                           \\
                                            TVL,      & VIT-LENS-2       & 7.00      &   {\cellcolor[HTML]{E8F1F9}57.16}       &      {\cellcolor[HTML]{E8F1F9}84.58}        &    {\cellcolor[HTML]{E8F1F9}84.69}      &       {\cellcolor[HTML]{E8F1F9}42.63}                            &    {\cellcolor[HTML]{E8F1F9}76.62} & \multicolumn{1}{c}{{\cellcolor[HTML]{E8F1F9}-}}                               \\
                                            TacQuad   & TLV-CoRe    &    0.30        &    {\cellcolor[HTML]{E8F1F9}\textbf{60.26}}      &   {\cellcolor[HTML]{E8F1F9}\textbf{86.53}}           &     {\cellcolor[HTML]{E8F1F9}\textbf{87.13}}     &     {\cellcolor[HTML]{E8F1F9}\textbf{47.25}}                              &    {\cellcolor[HTML]{E8F1F9}\textbf{76.87}} & \multicolumn{1}{c}{{\cellcolor[HTML]{E8F1F9}\textbf{79.35}}}                               \\ \bottomrule[1.2pt]
\end{tabular}%
\vspace{-0.3cm}
\caption{Performance (\%) comparison of different methods under three evaluation protocols: \colorbox[HTML]{F6E6E6}{intra-sensor evaluation}, \colorbox[HTML]{E2F0D9}{cross-sensor generalization}, and \colorbox[HTML]{E8F1F9}{multi-sensor generalization}. $\dag$ Note that we follow the default configuration of the TLV-Link repository, applying LoRA~\cite{hu2022lora} to fine-tune the tactile and vision encoders while keeping the language encoder frozen. $^\ast$For methods that cannot handle missing modalities, the tactile modality is used as a substitute for the missing vision modality in the Octopi dataset.}
\label{tab:main}
\vspace{-0.5cm}
\end{table*}

\begin{corollary}[Cross-Modal Performance]
\label{cor:cross_eval}
Let $\mathcal{A}_m^{\mathrm{task}_{m'}}$ be the accuracy of encoder $m$ on tasks of modality $m'$. Then under Theorem~\ref{thm:cross_modal},
\begin{equation}
\mathcal{A}_m^{\mathrm{task}_{m'}} 
\;\ge\;\mathcal{A}_{m'}^{\mathrm{task}_{m'}} \;-\; \Delta_{m,m'} \;-\; 
\frac{C}{r},
\end{equation}
where $\Delta_{m,m'}$ is a small modality-gap term and $C$ is a task-dependent constant. Hence, as the shared dimension $r$ grows, the cross-modal performance of modality $m$ approaches that of the best modality $m'$ up to a small gap.
\end{corollary}

These results imply that multi-level UBA alignment allows each modality to absorb useful information from the others, thereby improving its performance on cross-modal tasks. In practice, we indeed observe consistent performance gains in modal cross-evaluation (see Sec.~\ref{sec:times}).

\subsubsection{Batch-Size Stability}
\label{sec:stability}

We consider the effect of batch size in training. In a batch of size $N$, an anchor typically encounters $\mathbb{E}[N_{\mathrm{sim}}]=(N-1)p_{\mathrm{sim}}$ semantically similar negatives on average, where $p_{\mathrm{sim}}$ is the probability of semantic overlap. Thus there is a trade-off: \textbf{Small batches:} fewer negatives (weaker contrastive signal) but emphasize fine-grained distinctions. \textbf{Large batches:} many negatives (stronger alignment signal) but may bias toward coarser features.

\begin{theorem}[Batch-Size Stability]
\label{thm:batch_stability}
Let $\epsilon_N$ be the expected task error when using batch size $N$. Under sensor-invariance decoupling, the error gap between any two batch sizes satisfies
\begin{equation}
|\epsilon_N - \epsilon_{N'}|
\;\le\;
\frac{C_1}{1 + C_2\bigl(1 - \mathcal{I}(h^T; s)\bigr)},
\end{equation}
for constants $C_1,C_2>0$. As $\mathcal{I}(h^T; s)$ decreases (via the SAM), this bound shrinks, making performance less sensitive to the choice of $N$.
\end{theorem}

\begin{proposition}[Representation Enhancement]
\label{prop:feature_learning}
After UBA alignment, each representation satisfies
\begin{equation}
\begin{aligned}
&\mathcal{I}\!\bigl(h^m_{\mathrm{aligned}};Y\bigr)
\;\ge\;\mathcal{I}\!\bigl(h^m;Y\bigr)
\\
&+\;\max_{m'\neq m}\bigl[\mathcal{I}\!\bigl(h^{m'};Y\bigr)\;-\;\mathcal{I}\!\bigl(h^m;Y\bigr)\bigr]_+,
\end{aligned}
\end{equation}

so it gains any extra information present in the most informative other modality.
\end{proposition}

Theorem~\ref{thm:batch_stability} and Proposition~\ref{prop:feature_learning} together explain TLV-CoRe’s stability:
\begin{itemize}[leftmargin=*, itemsep=0pt, topsep=0pt]
    \item \textbf{Tactile Decoupling} ensures the tactile encoder focuses on invariant, task-relevant features, reducing spurious correlations and enabling stable learning even with small batches.
    \item \textbf{UBA Sharing} ensures each encoder learns fine-grained and global features by absorbing information from other modalities, making the learned representations robust to batch-size variations.
\end{itemize}

In summary, our theoretical analysis demonstrates that sensor-aware decoupling and shared adapters yield benefits in convergence, cross-modal transfer, and training stability. They align with our empirical performance findings (see Sec.~\ref{sec:batch1}).

Note that \textbf{we provide the proposed TLV-CoRe's empirical validation of convergence analysis and more detailed theoretical analysis in the Appendix}.

\section{Experiments}
\label{sec:experiment}

Under the RSS evaluation framework, our experiments primarily focus on tactile representations and adopt a comprehensive and consistent linear probing approach~\cite{cheng2025touch100k, feng2025anytouch, lei2024vit} to evaluate the quality of tactile representations from different methods across three evaluation protocols. The specific evaluation protocols include intra-sensor evaluation, cross-sensor generalization, and multi-sensor generalization, across two tasks—material property identification and robot grasping prediction.

\begin{table*}[ht]
\centering
\setlength{\tabcolsep}{2.5mm}
\begin{tabular}{@{}c|c|ccc|c|c|cc@{}}
\toprule[1.2pt]
\multirow{2.5}{*}{Training Data} & \multirow{2.5}{*}{Method} &          & TAG       &          & CIFAR-10          &    CIFAR-100    & ImgNetDogs      \\ \cmidrule(lr){3-5} \cmidrule(lr){6-6} \cmidrule(lr){7-7} \cmidrule(lr){8-8} 
                               &                         & Material & Roughness & Hardness & Image CLS & Image CLS & Image CLS \\  \midrule
\multirow{4}{*}{TAG}           & TLV-Link                &     {\cellcolor[HTML]{F6E6E6}52.91}     &    {\cellcolor[HTML]{F6E6E6}82.69}       &     {\cellcolor[HTML]{F6E6E6}85.17}     &             {\cellcolor[HTML]{F0F0F0}32.39}   &        {\cellcolor[HTML]{F0F0F0}10.88}        &           {\cellcolor[HTML]{F0F0F0}25.09}     \\
                               & AnyTouch                &   {\cellcolor[HTML]{F6E6E6}53.64}       & {\cellcolor[HTML]{F6E6E6}84.52}          &   {\cellcolor[HTML]{F6E6E6}85.19}       &             {\cellcolor[HTML]{F0F0F0}40.90}   &        {\cellcolor[HTML]{F0F0F0}18.18}        &          {\cellcolor[HTML]{F0F0F0}25.97}      \\
                               & VIT-LENS-2              &    {\cellcolor[HTML]{F6E6E6}53.32}      &  {\cellcolor[HTML]{F6E6E6}85.94}         &   {\cellcolor[HTML]{F6E6E6}86.13}       &             {\cellcolor[HTML]{F0F0F0}44.38}   &        {\cellcolor[HTML]{F0F0F0}19.97}        &     {\cellcolor[HTML]{F0F0F0}28.90}           \\
                               & TLV-CoRe                &    {\cellcolor[HTML]{F6E6E6}\textbf{53.86}}      &      {\cellcolor[HTML]{F6E6E6}\textbf{87.39}}     &   {\cellcolor[HTML]{F6E6E6}\textbf{88.62}}       &       {\cellcolor[HTML]{F0F0F0}\textbf{68.15}}         &             {\cellcolor[HTML]{F0F0F0}\textbf{34.22}}   &      {\cellcolor[HTML]{F0F0F0}\textbf{30.07}}         \\ \midrule
\multirow{4}{*}{Octopi$^\ast$}         & TLV-Link                &    {\cellcolor[HTML]{E2F0D9}48.72}      &     {\cellcolor[HTML]{E2F0D9}79.55}      &    {\cellcolor[HTML]{E2F0D9}81.97}      &             {\cellcolor[HTML]{F0F0F0}59.35}   &         {\cellcolor[HTML]{F0F0F0}31.22}       &     {\cellcolor[HTML]{F0F0F0}28.11}           \\
                               & AnyTouch                &    {\cellcolor[HTML]{E2F0D9}44.39}      &    {\cellcolor[HTML]{E2F0D9}\textbf{86.36}}       &  {\cellcolor[HTML]{E2F0D9}81.13}        &             {\cellcolor[HTML]{F0F0F0}38.64}   &         {\cellcolor[HTML]{F0F0F0}13.88}       &         {\cellcolor[HTML]{F0F0F0}24.11}      \\
                               & VIT-LENS-2              &    {\cellcolor[HTML]{E2F0D9}48.11}      &  {\cellcolor[HTML]{E2F0D9}82.02}         &   {\cellcolor[HTML]{E2F0D9}84.36}       &             {\cellcolor[HTML]{F0F0F0}43.23}   &         {\cellcolor[HTML]{F0F0F0}17.62}       &   {\cellcolor[HTML]{F0F0F0}26.75}             \\
                               & TLV-CoRe                &   {\cellcolor[HTML]{E2F0D9}\textbf{52.65}}       &  {\cellcolor[HTML]{E2F0D9}85.83}         &    {\cellcolor[HTML]{E2F0D9}\textbf{86.43}}      &  {\cellcolor[HTML]{F0F0F0}\textbf{70.46}}             &        {\cellcolor[HTML]{F0F0F0}\textbf{37.41}}        &    {\cellcolor[HTML]{F0F0F0}\textbf{28.25}}            \\ \midrule

\multirow{4}{*}{TacQuad}       & TLV-Link                &     {\cellcolor[HTML]{E8F1F9}54.15}     &      {\cellcolor[HTML]{E8F1F9}84.62}     &     {\cellcolor[HTML]{E8F1F9}85.97}     &  {\cellcolor[HTML]{F0F0F0}76.77}  &      {\cellcolor[HTML]{F0F0F0}52.39}   &            {\cellcolor[HTML]{F0F0F0}29.27}                 \\
                               & AnyTouch                &   {\cellcolor[HTML]{E8F1F9}50.37}       & {\cellcolor[HTML]{E8F1F9}84.50}          &    {\cellcolor[HTML]{E8F1F9}82.19}      &    {\cellcolor[HTML]{F0F0F0}50.67}   &     {\cellcolor[HTML]{F0F0F0}25.29}     &                 {\cellcolor[HTML]{F0F0F0}26.26}            \\
                               & VIT-LENS-2              &    {\cellcolor[HTML]{E8F1F9}51.29}      &  {\cellcolor[HTML]{E8F1F9}85.26}         &    {\cellcolor[HTML]{E8F1F9}84.72}      &     {\cellcolor[HTML]{F0F0F0}43.18}   &   {\cellcolor[HTML]{F0F0F0}18.76}  &  {\cellcolor[HTML]{F0F0F0}26.56}               \\
                               & TLV-CoRe                &     {\cellcolor[HTML]{E8F1F9}\textbf{56.52}}     &  {\cellcolor[HTML]{E8F1F9}\textbf{85.97}}         &    {\cellcolor[HTML]{E8F1F9}\textbf{86.47}}      &             {\cellcolor[HTML]{F0F0F0}\textbf{78.90}}   &         {\cellcolor[HTML]{F0F0F0}\textbf{52.70}}       &   {\cellcolor[HTML]{F0F0F0}\textbf{31.47}}    \\
                               \bottomrule[1.2pt]
\end{tabular}%

\vspace{-0.3cm}
\captionof{table}{Performance (\%) comparison of different methods in modal cross-evaluation. $^\ast$Since the visual input is replaced by tactile images, the vision encoder is equivalent to the tactile encoder.}\label{tab:synergy}
\vspace{-0.5cm}
\end{table*}

\section{Experimental Setup}
\label{subsec: Experimental}

\subsection{Real-World Tactile Datasets}
\label{sec:dataset}

We train various tactile representation learning methods on single-sensor datasets (Touch and Go (TAG)~\cite{yang2022touch}, SSVTP~\cite{kerr2022self}, TVL~\cite{fu2024tou}, and Octopi~\cite{yu2024octopi}) and a multi-sensor dataset (TacQuad~\cite{feng2025anytouch}). For material property identification, we select three test subsets from TAG, along with ObjectFolder 1.0 (OF 1.0)~\cite{gao2021objectfolder} and ObjectFolder 2.0 (OF 2.0)~\cite{gao2022objectfolder}, as downstream evaluation datasets. TAG includes three tactile classification tasks: material (20 classes), roughness (2 classes), and hardness (2 classes). OF 1.0 and OF 2.0 focus on material classification (7 classes). The two datasets splits follow the setup in~\cite{yang2024binding}. For robot grasping prediction, we evaluate on The Feeling of Success (Feel) dataset~\cite{calandra2017feeling}. Following~\cite{feng2025anytouch}, we use frames from the left and right tactile sensors during the grasping process as input to predict whether a grasp is successful. Based on~\cite{yang2024binding, cheng2025touch100k}, we split the Feel dataset by object into training, validation, and test sets with a ratio of 8:1:1. Table~\ref{tab:statics} presents the statistics of the aforementioned datasets. Notably, the language modality in TAG is generated by GPT-4o, and textual descriptions in TVL, SSVTP, and Octopi are extended, as detailed in~\cite{feng2025anytouch}.

\subsection{Tactile-Language-Vision Baselines}
\label{sec:baseline}

We compare the proposed TLV-CoRe method with state-of-the-art multimodal tactile approaches, including single-sensor methods TLV-Link~\cite{cheng2025touch100k} and VIT-LENS-2~\cite{lei2024vit}, as well as the multi-sensor method AnyTouch~\cite{feng2025anytouch}. Notably, all these methods use OpenCLIP-large~\cite{cherti2023reproducible} to initialize the tactile, language, and vision encoders, which allows for a fair comparison among different multimodal tactile approaches without the confounding factor of varying base models. Specifically, both the tactile and vision encoders adopt a 24-layer, 1024-dimensional Vision Transformer (ViT)~\cite{dosovitskiy2020image} with a patch size of 14, while the language encoder uses a 12-layer, 768-dimensional Transformer architecture~\cite{vaswani2017attention}. It's noted that since UniTouch~\cite{yang2024binding} is not fully open-source, we do not use it as a baseline in our work.

\subsection{Implementation Details}
\label{sec:implementation}

We train various tactile models for 12 epochs, except for VIT-LENS-2 which is trained for 80 epochs to ensure convergence, using the AdamW optimizer~\cite{loshchilov2017decoupled} with an initial learning rate of 2e-4 and momentum parameters $\beta_1, \beta_2 = 0.9, 0.98$ on two NVIDIA A800 GPUs. The first stage of AnyTouch is trained for 20 epochs. All models are trained with linear probing for 50 epochs. For both TLV-Link and AnyTouch, the LoRA rank is set to 16. VIT-LENS-2 adopts the VIT-LENS$_L$ architecture. For the proposed TLV-CoRe model, the shared projection layer dimension is set to 32. We apply UBA to all layers of the language encoder and introduce UBA to the tactile and vision encoders starting from layer $k=13$, ensuring one-to-one correspondence across the three modalities. In cross-sensor tactile decoupling, the hyperparameter $\lambda$ is set to 0.1. Unless otherwise specified, all models are trained with a batch size of 64, and other hyperparameters follow the original papers. Following previous work~\cite{yang2022touch, yang2024binding, cheng2025touch100k, feng2025anytouch, lei2024vit}, we use accuracy as the evaluation metric for classification tasks.

\subsection{Main Results $\Rightarrow$ \underline{R}obustness Analysis}
\label{sec:main}

We systematically compare various methods using three evaluation protocols, as shown in Table~\ref{tab:main}. Overall, among these CLIP-based approaches, the proposed \textbf{TLV-CoRe consistently outperforms existing methods with significantly fewer trainable parameters}. The following observations emerge:

\begin{itemize}[leftmargin=*, itemsep=0pt, topsep=0pt]
    \item{Across the three evaluation protocols, the overall performance ranks as: intra-sensor evaluation $>$ cross-sensor generalization $\approx$ multi-sensor generalization. This trend primarily stems from the fact that intra-sensor evaluation is an in-distribution test, where the patterns learned by the model closely match those in test set, leading to better generalization. In contrast, in the out-of-distribution settings of cross- and multi-sensor generalization, the model may struggle to interpret previously unseen features effectively, resulting in performance drops that cannot be easily mitigated by simply increasing the amount of training data.}
    \item{Although Octopi and TAG use the same type of GelSight sensors, models trained on Octopi generally underperform on the three TAG sub-tests. We attribute this to Octopi's lack of visual modality and limited data size, which leads to insufficient training.}
    \item{The performance differences among existing methods remain relatively minor.  VIT-LENS-2 performs better in some cases, possibly due to its larger number of trainable parameters. However, in general, these methods still lack robust performance across all three evaluation protocols.}
    \item{In contrast, the proposed TLV-CoRe consistently achieves leading results across all tasks, excelling particularly in material classification. While it may not always lead in roughness binary classification—likely due to the higher randomness of simpler tasks—it reliably ranks within the top two, with only a minimal margin from the best. This further validates the robustness and reliability of TLV-CoRe.}
\end{itemize}

\begin{table*}[ht]
\centering
\vspace{-0.1cm}
\begin{minipage}[t]{0.32\textwidth}
  \centering
  \setlength{\tabcolsep}{0.7mm}
  \begin{tabular}{@{}c|ccc@{}}
  \toprule[1.2pt]
  $\lambda_{DL}$ & Material           & Roughness          & Hardness             \\ \midrule
  0.01   &  56.30         &  84.78         &    85.83       \\
  0.05   &  57.64         & 85.13 &    86.25       \\
  0.1   & 58.37 & \textbf{86.80}     & \textbf{87.52} \\
  0.5   &  \textbf{58.73}         & 86.40          &    87.12      \\
  1.0   &  58.64         &    86.71       &   86.17        \\ \bottomrule[1.2pt]
  \end{tabular}%
\end{minipage}
\hfill
\begin{minipage}[t]{0.32\textwidth}
  \centering
  \setlength{\tabcolsep}{0.7mm}
  \begin{tabular}{@{}c|ccc@{}}
  \toprule[1.2pt]
  $N_{UBA}$ & Material           & Roughness          & Hardness             \\ \midrule
  4   &  56.64         &   84.93        &    86.15       \\
  6  &   57.11        &   86.15        &     86.53      \\
  8  & 57.47  & \textbf{86.93} & 87.26  \\
  10  &  57.99        &  86.51         &    87.49       \\
  12 & \textbf{58.37}          & 86.80          & \textbf{87.52}          \\ \bottomrule[1.2pt]
  \end{tabular}%
\end{minipage}
\hfill
\begin{minipage}[t]{0.32\textwidth}
  \centering
  \setlength{\tabcolsep}{0.7mm}
  \begin{tabular}{@{}c|ccc@{}}
  \toprule[1.2pt]
  $d_{shared}$ & Material  & Roughness & Hardness    \\ \midrule
  8   &   55.92        &   84.49        &  86.36         \\
  16   &  57.26         &  85.31 &     86.54      \\
  32    & 58.37          & \textbf{86.80}          & 87.52 \\
  64    &  \textbf{58.52}         &  86.27         &   87.05        \\
  128     & 57.93 &     86.47      &   \textbf{87.61}        \\ \bottomrule[1.2pt]
  \end{tabular}%
\end{minipage}
\vspace{-0.2cm}
\caption{Ablation experiments on hyperparameters.}  
\label{tab:ablation}
\vspace{-0.3cm}
\end{table*}

\subsection{Tactile $\times$ Vision $\Rightarrow$ \underline{S}ynergy Analysis}
\label{sec:times}

The \textbf{goal of cross-modal alignment is to achieve information complementarity between modalities, rather than sacrificing the representation quality of each modality}~\cite{wang2022deep, dufumier2024align}. In other words, the aim is to enable synergy across modalities. To evaluate how well different methods achieve this synergy between tactile and vision modalities, we design a modal cross-evaluation. We select models trained on the TAG, Octopi, and TacQuad datasets in Sec.~\ref{sec:main} as the evaluation subjects. For tactile tasks, we choose three subtasks from the TAG dataset and evaluate them using the corresponding models’ vision encoders. For vision tasks, we use three image classification datasets—CIFAR-10, CIFAR-100~\cite{krizhevsky2009learning}, and ImgNetDogs~\cite{raduly2018dog}—and evaluate them with the tactile encoders. Table~\ref{tab:synergy} shows the following observations: \texttt{(1)} In tactile tasks, vision encoders generally perform worse than tactile encoders, which is expected given their modality focus. \texttt{(2)} In vision tasks, most methods demonstrate limited performance, possibly due to the setting of only 50 linear probing training epochs, which may indicate that a model with only a learnable classification head may not have sufficiently learned the new visual information. \texttt{(3)} Notably, the proposed TLV-CoRe achieves the best performance across both types of tasks. We attribute this to the introduction of the \textit{UBA} module, which bridges the modality branches via parameter sharing. This facilitates a unified feature space, enabling smooth gradient flow and efficient information transfer and complementarity across modalities, while also accelerating convergence. These results further validate the analysis in Sec.~\ref{sec:conv1}.

\begin{figure}[t]
    \centering
    \vspace{-0.25cm}
    \includegraphics[width=\columnwidth]{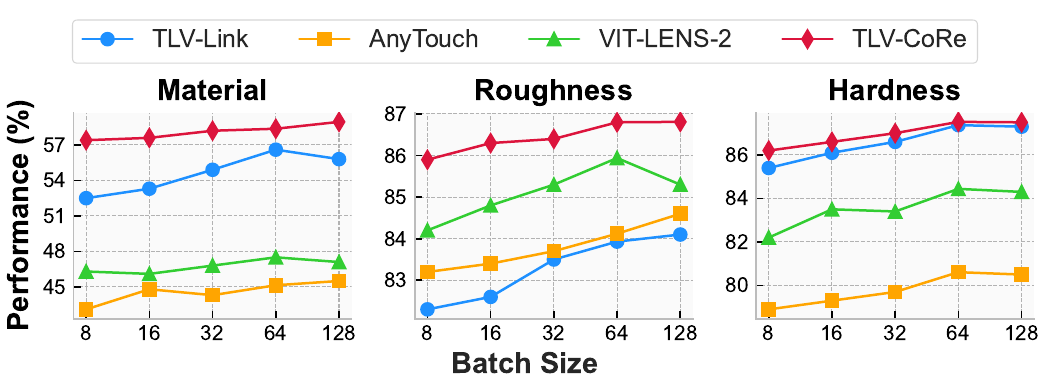}
    \vspace{-0.7cm}
    \caption{Performance (\%) comparison of different methods across various batch sizes.}\label{fig:batch}
    \vspace{-0.5cm}
\end{figure}

\subsection{Different Batch Sizes $\Rightarrow$ \underline{S}tability Analysis}
\label{sec:batch1}

In CLIP-based contrastive methods, different batch sizes lead to significant variations in the number of negative samples. As illustrated in Fig.\ref{fig:batch}, batch size significantly affects the performance of various methods trained by TacQuad dataset, evaluated across three subsets of TAG dataset. Notably, larger batch sizes tend to yield better performance, which aligns with observations from previous studies~\cite{chen2020simple, kerr2022self}. However, we also observe that when the batch size reaches 128, the performance of TLV-Link and VIT-LENS-2 no longer improves and may even decline. This is because a larger batch contains more data from different sensors, making it harder for single-sensor methods to handle the increased heterogeneity in tactile data. Overall, the proposed TLV-CoRe exhibits a smoother performance curve and demonstrates superior stability. This observation is consistent with the analysis presented in Sec.~\ref{sec:stability}.

\subsection{Ablation Study}

\begin{figure}[t!]
    \centering
    \vspace{-0.5cm}
    \includegraphics[width=0.8\columnwidth]{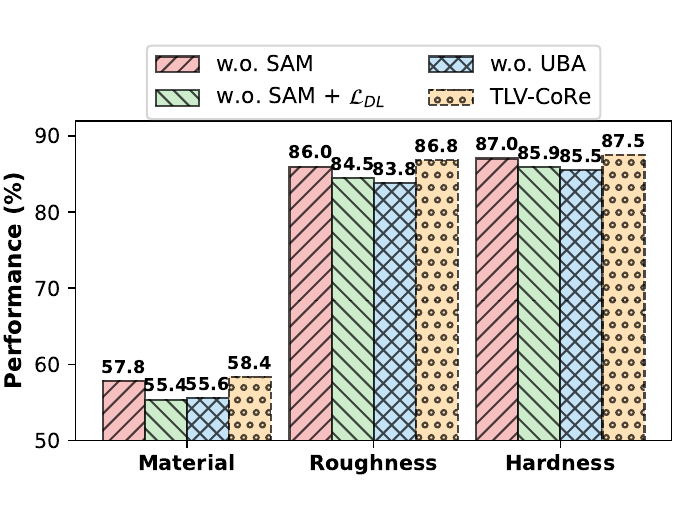}
    \vspace{-0.8cm}
    \caption{Ablation experiments on various components.}\label{fig:ablation}
    \vspace{-0.5cm}
\end{figure}

We perform an ablation study on the components of TLV-CoRe trained by TacQuad. As shown in Fig.~\ref{fig:ablation}, removing the SAM leads to a performance drop, and eliminating the decoupled loss $\mathcal{L}_{DL}$ causes an even greater decline, highlighting the crucial role of decoupled learning in handling multi-sensor data. Additionally, removing the UBA module results in the worst performance, underscoring its critical role in the overall architecture.

Additionally, we further investigate the impact of hyperparameters on TLV-CoRe trained by TacQuad dataset: the decoupled coefficient $\lambda_{DL}$, the number of UBA $N_{UBA}$, and the dimension of UBA shared layer $d_{shared}$. As shown in Table~\ref{tab:ablation}, both $\lambda_{DL}$ and $d_{shared}$ involve a trade-off between performance and generalization: a low $\lambda_{DL}$ weakens the decoupled effect, while a high value suppresses contrastive learning; a small $d_{shared}$ leads to insufficient sharing, whereas a large one tends to cause overfitting. In contrast, increasing $N_{UBA}$ generally improves performance, which indicates that it's necessary to fully promote collaboration and communication between different modalities.

\begin{table}[ht]
\centering
\setlength{\tabcolsep}{0.5mm}
\begin{tabular}{@{}c|c|ccc@{}}
\toprule[1.2pt]
Sensor Data                 & Method   & Material & Roughness & Hardness \\ \midrule
\multirow{2}{*}{GelSight}       & AnyTouch & 42.16    & 80.73     & 80.36    \\
                                & TLV-CoRe & 54.41    & 83.74     & 84.53    \\ \midrule
\multirow{2}{*}{DIGIT}          & AnyTouch & 42.75    & 79.49     & 79.24    \\
                                & TLV-CoRe & 55.23    & 80.16     & 82.80    \\ \midrule
\multirow{2}{*}{GelSight, DIGIT} & AnyTouch & 43.83    & 79.26     & 78.25    \\
                                & TLV-CoRe & 55.59    & 82.35     & 85.47    \\ \bottomrule[1.2pt]
\end{tabular}%
\vspace{-0.2cm}
\captionof{table}{Performance (\%) comparison of multi-sensor AnyTouch and TLV-CoRe across various sensor data.}
\label{tab:sensor}
\end{table}

To investigate the impact of tactile images with similar styles but from different sensors on model performance, we conduct experiments using the GelSight and DIGIT data from TacQuad dataset, as they exhibit stylistic similarity. Specifically, we randomly sample 5,000 instances from each dataset and additionally sample 2,500 instances from each to form a combined GelSight + DIGIT dataset. Due to the limited data size, we set the batch size to 8. Table~\ref{tab:sensor} compares the performance of the multi-sensor methods AnyTouch and TLV-CoRe. Results show that AnyTouch suffers a performance drop when shifting from single-sensor data (GelSight or DIGIT) to the mixed dataset, indicating its inability to handle style-consistent sensor differences, which leads to degraded multi-sensor representations. In contrast, TLV-CoRe mitigates this issue through decoupled learning and demonstrates more stable performance.

\section{Conclusion}
\label{sec:conclusion}
In this paper, we present TLV-CoRe, a collaborative representation learning method for tactile, language, and vision modalities. TLV-CoRe introduces a \textit{Sensor-Aware Modulator} to unify tactile representations across various sensors, employs tactile-irrelevant decoupled learning to disentangle tactile-irrelevant features, and incorporates a \textit{Unified Bridging Adapter} to enhance tri-modal interaction. To support fair evaluation, we propose the RSS framework. Experimental results show that TLV-CoRe achieves strong performance.

\section{Limitations}
\label{sec:limitation}

Our work focuses on aligning tactile, language, and vision modalities to explore the relatively under-researched area of tactile representation, with a particular emphasis on CLIP-based tactile representation learning. We introduce a fair evaluation framework called RSS for tactile tasks. Our experiments are conducted using real-world tactile data. While these datasets are representative and diverse, they may still fall short of fully capturing the complexity of real-world scenarios. A key next step is to conduct evaluations on real robotic systems to gain deeper insights into the model's performance in real-time tasks. Furthermore, expanding the evaluation scope to include more complex manipulation tasks will help provide a more comprehensive understanding of the model’s capabilities and limitations. Notably, our experiments are based on data collected from real-world tasks, which enhances the practical relevance and significance of our findings.

\section*{Acknowledgments}
This research was supported by grants from the ``Pioneer'' and ``Leading Goose'' R\&D Program of Zhejiang (2025C02022) and National Natural Science Foundation of China (No.62307032). Additionally, this work was partially supported by the UK Engineering and Physical Sciences Research Council (EPSRC) [EP/W524694/1].

\bibliography{aaai2026}

\clearpage
\appendix

\section{Convergence Analysis}
\label{subsec:empirical_convergence}

\begin{figure}[t]
    \centering
    \includegraphics[width=0.95\linewidth]{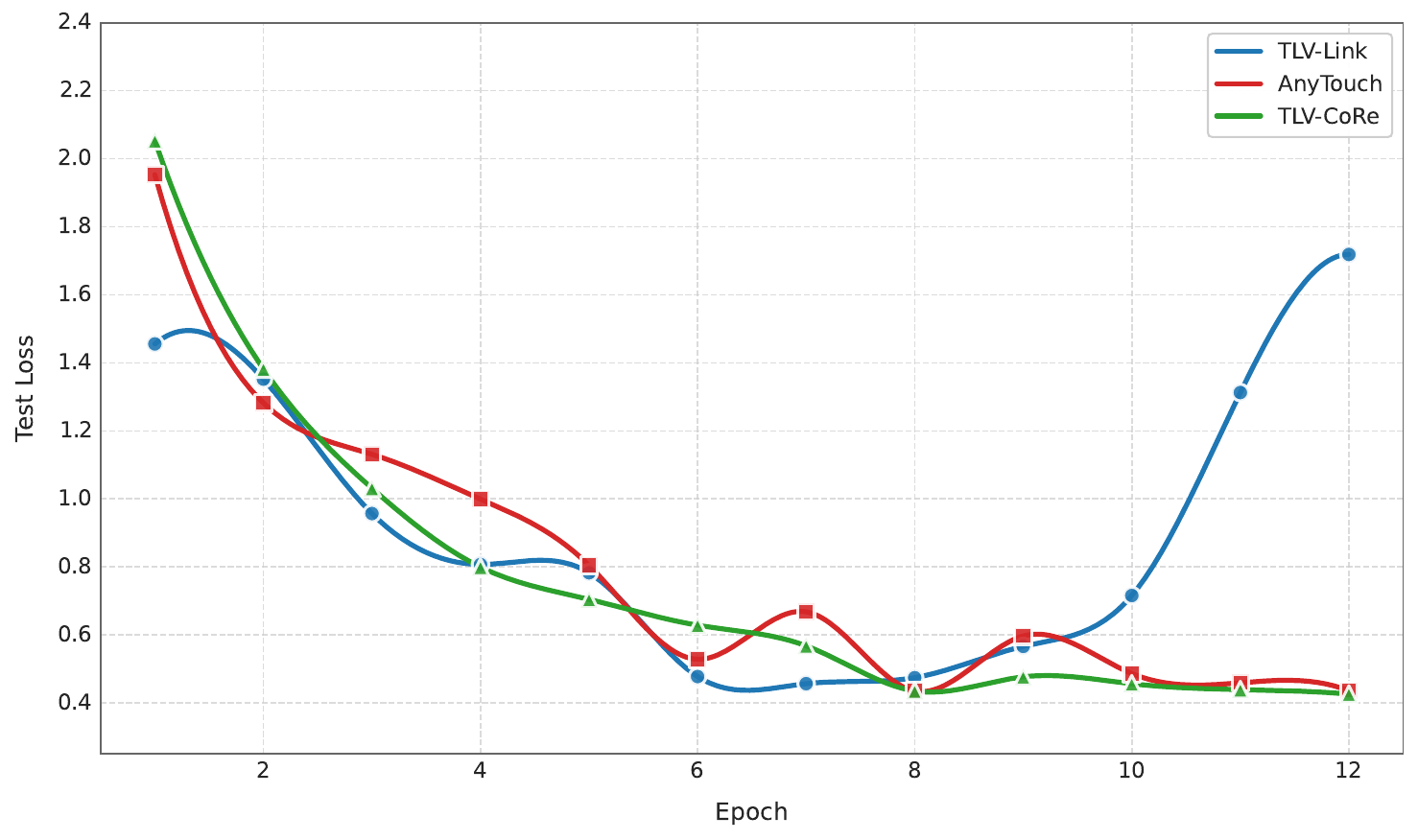}
    \vspace{-0.2cm}
    \caption{Convergence comparison of TLV-CoRe versus state-of-the-art baselines on the TAG dataset. Each line represents the test loss. TLV-CoRe (green) exhibits faster convergence and more stability compared to TLV-Link (blue) and AnyTouch (red), empirically validating our theoretical analysis in Theorem 3.1 and Proposition 3.3. Notably, while TLV-Link suffers from instability after epoch 8, TLV-CoRe maintains stable performance throughout training.}
    \label{fig:convergence}
\end{figure}

To validate our theoretical analysis of convergence properties in Theorem 3.1, we compare the training performance of TLV-CoRe against two state-of-the-art approaches: TLV-Link~\cite{cheng2025touch100k} and AnyTouch~\cite{feng2025anytouch}. Fig.~\ref{fig:convergence} shows the test loss trajectories over 12 epochs on the TAG dataset. The results provide strong empirical support for our theoretical claims. First, TLV-CoRe exhibits the fastest convergence rate among all methods, achieving a stable loss value by epoch 8, while other approaches continue to fluctuate. Second, unlike TLV-Link which shows instability after epoch 8 with increasing loss, TLV-CoRe maintains consistent performance. The combined effect of these properties enables TLV-CoRe to achieve both lower final test loss and more reliable optimization behavior, creating a foundation for robust downstream performance in this datasets and others.

\section{Detailed Theoretical Analysis}
\label{subsec: Theoretical Supp}
We provide detailed proofs of the theoretical results stated in Sec. 3.4. Before proceeding, recall the standard assumptions (Lipschitz gradient, PL condition, bounded variance) from Sec. 3.4.

\begin{proof}[Proof of Theorem 3.1]
Under the Lipschitz and PL assumptions, the progress of SGD satisfies the standard linear convergence analysis. In particular, consider an SGD step $\Theta_{t+1}=\Theta_t - \eta\nabla\mathcal{L}_{\mathcal{B}}(\Theta_t)$ where $\nabla\mathcal{L}_{\mathcal{B}}$ is the stochastic gradient on a mini-batch $\mathcal{B}$.  Taking expectation and using 
$\mathbb{E}\bigl[\|\nabla \mathcal{L}_{\mathcal{B}}(\Theta_t)\|^2\bigr] = \|\nabla \mathcal{L}(\Theta_t)\|^2 + \Var(\nabla \mathcal{L}_{\mathcal{B}})$ 
along with the bounded variance assumption, we have

\[
\begin{alignedat}{1}
&\mathbb{E}\|\Theta_{t+1}-\Theta^*\|^2 \\[2pt]
&= \mathbb{E}\bigl\|\Theta_t-\Theta^*-\eta\nabla\mathcal{L}_{\mathcal{B}}(\Theta_t)\bigr\|^2 \\[2pt]
&= \mathbb{E}\|\Theta_t-\Theta^*\|^2 \\ 
&\quad - 2\eta\,\mathbb{E}\langle\nabla\mathcal{L}(\Theta_t),\Theta_t-\Theta^*\rangle \\ 
&\quad + \eta^2\mathbb{E}\|\nabla\mathcal{L}_{\mathcal{B}}(\Theta_t)\|^2 \\[2pt]
&\le \mathbb{E}\|\Theta_t-\Theta^*\|^2 \\ 
&\quad - 2\eta\,\langle\nabla\mathcal{L}(\Theta_t),\Theta_t-\Theta^*\rangle \\ 
&\quad + \eta^2\bigl(\|\nabla\mathcal{L}(\Theta_t)\|^2 + \sigma^2\bigr).
\end{alignedat}
\]

The PL condition implies $\|\nabla\mathcal{L}(\Theta_t)\|^2\ge 2\mu\bigl(\mathcal{L}(\Theta_t)-\mathcal{L}(\Theta^*)\bigr)$ and 
$\mathcal{L}(\Theta_t)-\mathcal{L}(\Theta^*) \ge \tfrac{\mu}{2}\|\Theta_t-\Theta^*\|^2$. Hence 
$\|\nabla\mathcal{L}(\Theta_t)\|^2 \ge \mu^2\|\Theta_t-\Theta^*\|^2$.  Moreover, local strong convexity gives 
$\langle \nabla\mathcal{L}(\Theta_t), \Theta_t-\Theta^*\rangle \ge \mu \|\Theta_t-\Theta^*\|^2$.  Substituting these bounds,
\[
\begin{alignedat}{1}
&\mathbb{E}\|\Theta_{t+1}-\Theta^*\|^2 \\[2pt]
&\le \mathbb{E}\|\Theta_t-\Theta^*\|^2 \\ 
&\quad - 2\eta\mu\,\mathbb{E}\|\Theta_t-\Theta^*\|^2 \\ 
&\quad + \eta^2\bigl(\mu^2\,\mathbb{E}\|\Theta_t-\Theta^*\|^2 + \sigma^2\bigr) \\[2pt]
&= (1 - 2\eta\mu + \eta^2\mu^2)\,\mathbb{E}\|\Theta_t-\Theta^*\|^2 \\ 
&\quad + \eta^2\sigma^2.
\end{alignedat}
\]

Choose $\eta<1/\mu$.  Then $1 - 2\eta\mu + \eta^2\mu^2 \le 1 - \eta\mu$, so 
\[
\mathbb{E}\|\Theta_{t+1}-\Theta^*\|^2 \;\le\; (1 - \eta\mu)\,\mathbb{E}\|\Theta_t-\Theta^*\|^2 + \eta^2\sigma^2.
\]
Unrolling this recurrence yields 
\[
\begin{alignedat}{1}
&\mathbb{E}\|\Theta_t-\Theta^*\|^2 \\[2pt]
&\;\le\; (1-\eta\mu)^t\,\|\Theta_0-\Theta^*\|^2 \\ 
&\quad + \frac{\eta\sigma^2}{\mu}\bigl(1 - (1-\eta\mu)^t\bigr).
\end{alignedat}
\]

Taking $t\to\infty$ gives the asymptotic bound 
\[
\mathbb{E}\|\Theta_t-\Theta^*\|^2 \;\le\; (1 - \eta\mu)^t \|\Theta_0-\Theta^*\|^2 + \frac{\eta\sigma^2}{\mu}.
\]
Finally, the shared adapter $W_{\mathsf{sh}}$ improves the effective conditioning of the problem.  In effect, its condition number $\kappa(W_{\mathsf{sh}})$ scales the curvature of the shared subspace, which can be shown to replace $\mu$ by $\mu/\beta$ with $\beta=1/(1+\kappa(W_{\mathsf{sh}}))$.  Substituting $\mu\to\mu/\beta$ in the above bound yields the stated result.
\end{proof}

\begin{proof}[Proof of Lemma 3.2]
Let $g(\Theta)=\nabla\mathcal{L}(\Theta)$ denote the full gradient of the loss, which depends on the tactile feature $h^T$ and sensor label $s$.  Decompose the total variance by conditioning on $s$:
\[
\Var(g) \;=\; \mathbb{E}_s\bigl[\Var(g\mid s)\bigr] + \Var_s\bigl[\mathbb{E}(g\mid s)\bigr].
\]
The term $\mathbb{E}_s[\Var(g\mid s)]$ is the variance over mini-batches for a fixed sensor and is bounded by $\sigma_0^2$.  The term $\Var_s[\mathbb{E}(g\mid s)]$ measures how the mean gradient varies with $s$.  As $\mathcal{I}(h^T;s)$ decreases, the distribution of $h^T$ given $s$ approaches the marginal, forcing the conditional means $\mathbb{E}(g\mid s)$ to cluster around the global mean $\mathbb{E}(g)$.  By Pinsker’s inequality (or a Taylor expansion of KL), one can show 
\[
\bigl\|\mathbb{E}(g\mid s) - \mathbb{E}(g)\bigr\|^2 \;\le\; C\,\mathcal{I}(h^T;s)
\]
for some constant $C$.  Averaging over $s$ then gives $\Var_s[\mathbb{E}(g\mid s)] \le C\,\mathcal{I}(h^T;s)$.  Combining these yields
\[
\Var(g) \;\le\; \sigma_0^2 - \gamma\,\mathcal{I}(h^T;s)
\]
for some $\gamma>0$, as claimed.  Thus reducing $\mathcal{I}(h^T;s)$ lowers the stochastic gradient variance.
\end{proof}

\begin{proof}[Proof of Proposition 3.3]
This follows directly from Lemma 3.2.  As SAM removes sensor-specific information, $\mathcal{I}(h^T;s)\to\varepsilon$.  Lemma 3.2 then gives $\Var(\nabla\mathcal{L}) \le \sigma_0^2 - \gamma\varepsilon$.  For small $\varepsilon$ we may rewrite $\gamma\varepsilon = \gamma'(1-\varepsilon)$ for a constant $\gamma'>0$, yielding $\Var(\nabla\mathcal{L}) \le \sigma_0^2 - \gamma'(1-\varepsilon)$, as stated. 
\end{proof}

\begin{proof}[Proof of Theorem 3.4]
Under Assumption 3.4, decompose the mutual information for modalities $m$ and $m'$ into shared and unique parts:
\[
\mathcal{I}(h^m;Y) = S + U_m,\qquad
\mathcal{I}(h^{m'};Y) = S + U_{m'},
\]
where $S$ is the information common to both modalities and $U_m,U_{m'}$ are the unique components.  Without loss of generality assume $\mathcal{I}(h^{m'};Y)\ge \mathcal{I}(h^m;Y)$, so $U_{m'}\ge U_m$.  The aligned representation is $h^m_{\mathrm{aligned}} = h^m + \Delta h^m$, and by the chain rule of mutual information:
\[
\begin{alignedat}{1}
&\mathcal{I}(h^m_{\mathrm{aligned}};Y)
  = \mathcal{I}\!\bigl(h^m+\Delta h^m;Y\bigr) \\[2pt]
&\ge \mathcal{I}(h^m;Y) \\
&\quad + \mathcal{I}\!\bigl(\Delta h^m;Y \mid h^m\bigr).
\end{alignedat}
\]

The additional term $\mathcal{I}(\Delta h^m;Y\mid h^m)$ represents new information about $Y$ contributed by $\Delta h^m$.  Since $\Delta h^m$ comes from the shared $r$-dimensional bottleneck, it can carry at most $\min\{r,\;U_{m'}-U_m\} = \min\{r,\;\mathcal{I}(h^{m'};Y)-\mathcal{I}(h^m;Y)\}$ new bits about $Y$ from modality $m'$.  Allowing for inefficiency, we include a factor $\alpha\in(0,1)$ to write
\[
\begin{aligned}
&\mathcal{I}\!\bigl(h^m_{\mathrm{aligned}};Y\bigr)
\;\ge\;\mathcal{I}\!\bigl(h^m;Y\bigr)
\\
&+\;\alpha\,\min\!\bigl\{\,r,\;\mathcal{I}\!\bigl(h^{m'};Y\bigr)-\mathcal{I}\!\bigl(h^m;Y\bigr)\bigr\},
\end{aligned}
\]
as claimed.
\end{proof}

\begin{proof}[Proof of Corollary 3.5]
From Theorem 3.4, $\mathcal{I}(h^m_{\mathrm{aligned}};Y)$ is within $O(1/r)$ of $\mathcal{I}(h^{m'};Y)$.  By Fano’s inequality, any classifier built from a representation $h$ has error $\epsilon(h)$ satisfying
\[
H(\epsilon(h)) + \epsilon(h)\log(|\mathcal{Y}|-1) \;\ge\; H(Y) - \mathcal{I}(h;Y).
\]
Applying this to $h^m_{\mathrm{aligned}}$ and $h^{m'}$, and noting that their mutual informations differ by only $O(1/r)$, it follows that $\epsilon_m \le \epsilon_{m'} + O(1/r) + \Delta_{m,m'}$.  Converting to accuracy ($\mathcal{A}=1-\epsilon$) gives 
\[
\mathcal{A}_m^{\mathrm{task}_{m'}} 
\;\ge\; \mathcal{A}_{m'}^{\mathrm{task}_{m'}} \;-\; \Delta_{m,m'} \;-\; O(1/r),
\]
which is the stated bound.
\end{proof}

\begin{proof}[Proof of Theorem 3.6]
The effect of batch size on contrastive loss can be understood via the noise in SGD updates.  Empirically, one can model the error as 
\[
\epsilon_N \;\approx\; \epsilon_\infty + \frac{A}{N},
\]
where $A$ depends on gradient variance.  Since SAM reduces variance according to $\Var(\nabla\mathcal{L})\approx \sigma_0^2-\gamma\mathcal{I}(h^T;s)$, the difference $\epsilon_N-\epsilon_{N'}$ becomes smaller as $\mathcal{I}(h^T;s)$ decreases.  More formally, comparing two such models for $N$ and $N'$ and using the bound on $\Var(\nabla\mathcal{L})$ yields
\[
|\epsilon_N - \epsilon_{N'}| \;\le\; \frac{C_1}{1 + C_2\bigl(1 - \mathcal{I}(h^T; s)\bigr)},
\]
for some constants $C_1,C_2>0$.  As $\mathcal{I}(h^T;s)$ decreases, the denominator grows and the gap shrinks, proving the result.
\end{proof}

\begin{proof}[Proof of Proposition 3.7]
Let $m^* = \arg\max_{m'\neq m}\mathcal{I}(h^{m'};Y)$ be the modality with the most information.  By Theorem 3.4, aligning $h^m$ with $h^{m^*}$ yields 
\[
\begin{alignedat}{1}
&\mathcal{I}(h^m_{\mathrm{aligned}};Y) \\[2pt]
&\;\ge\; \mathcal{I}(h^m;Y) \\ 
&\quad + \alpha\,\min\Bigl\{\,r,\; \mathcal{I}(h^{m^*};Y) - \mathcal{I}(h^m;Y)\Bigr\}.
\end{alignedat}
\]
If the shared dimension $r$ is large (or by iterating multiple alignments), we can take $\alpha\to1$ and $r\ge \mathcal{I}(h^{m^*};Y)-\mathcal{I}(h^m;Y)$, yielding 
$\mathcal{I}(h^m_{\mathrm{aligned}};Y)\ge \mathcal{I}(h^{m^*};Y)$.  More generally, even without full saturation, this implies
\[
\begin{aligned}
&\mathcal{I}\!\bigl(h^m_{\mathrm{aligned}};Y\bigr)
\;\ge\;\mathcal{I}\!\bigl(h^m;Y\bigr)
\\
&+\;\max_{m'\neq m}\bigl[\mathcal{I}\!\bigl(h^{m'};Y\bigr)\;-\;\mathcal{I}\!\bigl(h^m;Y\bigr)\bigr]_+,
\end{aligned}
\]
since the alignment can capture the largest information gap from the other modalities.  This completes the proof.
\end{proof}

\end{document}